# Evidence Combination and Reasoning and Its Applications to Real-World Problem Solving

L.W. Chang
Naval Research Laboratory
Code 5513
Washington, DC 20375-5000

R.L. Kashyap
School of Electrical Engineering
Purdue University
West Lafayette, IN 47907

*Abstract-* In this paper a new mathematical procedure is presented for combining different pieces of evidence which are represented in the interval form to reflect our knowledge about the truth of a hypothesis. Evidences may be correlated to each other (*dependent evidences*) or conflicting in supports (*conflicting evidences*). First, assuming independent evidences, we propose a methodology to construct combination rules which obey a set of essential properties. The method is based on a geometric model. We compare results obtained from Dempster-Shafer's rule and the proposed combination rules with both conflicting and non-conflicting data and show that the values generated by proposed combining rules are in tune with our intuition in both cases. Secondly, in the case that evidences are known to be dependent, we consider extensions of the rules derived for handling conflicting evidence. The performance of proposed rules are shown by different examples. The results show that the proposed rules reasonably make decision under dependent evidences.

## 1. Introduction

Evidence combination which pools different rational agents' judgments is essential to many practical applications, such as classification, diagnosis and radar system. In almost all these applications, the key step in reasoning process is to generate descriptions of target hypothesis and to aggregate these descriptions to form conclusion ([Booker, 1988]). Descriptions are transformed into numerical values or *beliefs* which are resulted from compatibility calculation, i.e., matching raw measures with models of known classes. The reasoning process is considerably more difficult when precise measures cannot be obtained. Under such circumstances, the interval-valued belief representation is used to describe real situations. The upper (lower) bound indicates the optimistic (pessimistic) estimate about the truth of target hypothesis and the width of interval reflects the measure of imprecision.

In evidence combination, theoretical framework of probability, Dempster-Shafer's theory of evidence and possibility theory have been proposed to discuss the problem. The relation between measures and target hypothesis can be described by network model which consists of evidence nodes and hypothesis node with links indicating conjunction of evidences mapped down to the belief of hypothesis. Thus, the assessment of high order joint conditional probabilities is required to determine the belief of target hypothesis. In the absence of this information, the Bayesian decision rule is used to combine each individual conditional probability. Interval combination with Bayesian decision rule has been discussed in [Cheng & Kashyap, 1988]. Dempster-Shafer's rule of combination ([Dempster, 1967][Shafer, 1986]) has received great attention recently. The implementation of this rule is carried out by normalization and multiplication of weights of support, conflict and ignorance where evidence is assumed to be independent and distinct. The combination technique applied in possibility theory extends the concept of the rule where normalization is not needed [Zadeh, 1986].

How do we evaluate different combination rules? The first criterion is that they should possess some basic properties like associativity, commuta-tivity etc. We can construct many rules having these properties including old favorites like Bayes or D-S rules. The key point in which these rules can be compared is the principles by which they combine conflicting evidence. For instance, does the final result reflect the fact that the components were in conflict? To illustrate this, take two evidences represented by intervals [0.15, 0.25] and [0.8, 0.9]. Clearly they are in conflict because the first interval [0.15, 0.25] is a subset of [0, 0.5] and indicates that it is highly unlikely that the hypothesis is true. On the contrary, the second evidence [0.8, 0.9] lies in the interval [0.5, 1] and indicates that hypothesis is likely to be true. The question is how can we develop a combination rule which takes into account the fact that the two components are conflicting. If we use the D-S rule, the result is the interval [0.56, 0.58]. The narrowness of the interval is striking and the entire interval lies in [0.5, 1]. The narrowness of interval indicates that the evidence is decisive, which in the case is not. This feature is there in D-S theory, by default, interval is always less than the width of intervals of the component evidences, regardless of whether the intervals are in conflict or not. This feature is clearly undesirable. On the other hand, interval resulted from using Bayesian decision rule may provide very conservative results even when a pair of evidence are *not* in conflict.

Furthermore most combining rules like D-S and Bayesian decision rule assume that the evidences are statistically independent. However, in practice, it is difficult to test the condition of independency. Sometimes we know that the two evidences are dependent, i.e., the two experts who arrived at the intervals used the same raw data. Then combining the evidences by D-S rules is roughly equivalent to using the same evidence twice. We have to consider the modification of the decision rule to handle dependent evidences.

Several methods have been proposed to achieve satisfactory results of combining conflicting or dependent evidence. In [Dubois & Prade, 1988], [Yager, 1987] and [Hau & Kashyap, 1987], approaches for handling conflicting evidence are to reallocate the weights associated with conflicting components to the ignorance or their disjunctions. In dealing with dependent evidence, the method suggested in [Blockley & Baldwin, 1987] is to control weights assigned to conjunctive terms. In [Dubois & Prade, 1986] and [Dubois & Prade, 1988] the minimum specificity principle is employed to specify joint assignments which cannot be obtained through D-S's rule. These combination rules provide more flexibility in belief combination. However, unlike D-S's rule, they do not obey associativity in general. In the absence of this property, these rules are not able to handle evidence combination step by step which is required in all real-time systems.

In this paper, we present a framework to deal with belief combination of various types of evidence and also ensure the combination satisfying some fundamental properties. Section 2 will deal with the basic axioms that should be satisfied by all rules. Section 3 will discuss the problem of conflicting evidence and how it can be handled. Section 4 discusses our approach to developing decision rules which obey all the necessary axioms and handles the conflicting evidences systematically. Section 6 handles the dependency problems. Section 8 gives the conclusions.

## 2. Necessary Properties or Axioms

Every evidence discussed here is represented by a numerical interval, say [a,b]. We will state this as a definition.

*Definition* (evidence). An evidence $e$ regarding a hypothesis $H$ is represented by a numerical interval [a, b] for the conditional belief $p(H|e)$, i.e.,

$$[a,b] \in S \iff 0 \leq a \leq p(H|e) \leq b \leq 1 \qquad (1)$$

(The $p(H|e)$ is not traditional probability, but it

obeys properties which will be explained later.) Thus if we represent [a,b] as a vector in a two dimensional coordinates system, S will be a triangle as in Figure 1 ([Rollinger, 1983]).

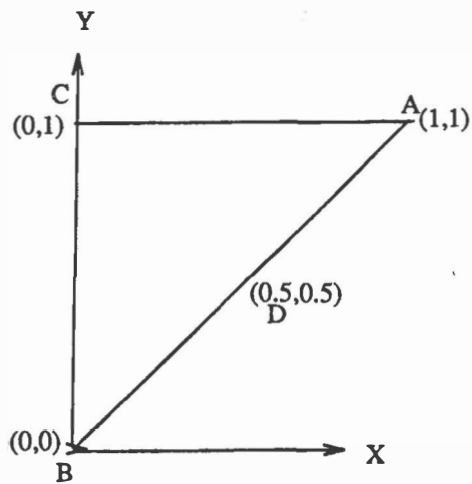

**Figure 1** Triangle Region

We will indicate a combination rule by the operator *. Thus the result of combining two intervals [a,b] and [c,d] is indicated by [a,b]*[c,d].

The first necessary property is closure, labelled (A1), so that result obeys (1)

**A1.** Closure

If $[a,b] \in S$ and $[c,d] \in S$,

then $[a,b]*[c,d] \in S$

where S is the triangle region shown in Figure 1. The next axiom (A2), commutativity states the result of combining two evidences cannot depend on the order in which they are combined, i.e., there is no ordering among evidences.

**A2.** Commutativity

$$[a,b] * [c,d] = [c,d] * [a,b]$$

The next axiom associativity deals with the requirement that when we have several (more than two) evidences, combine them pairwise, the final result is independent of the order in which they are combined.

**A3.** Associativity

$$([a,b] * [c,d]) * [e,f] = [a,b] * ([c,d] * [e,f])$$

The next axiom (A4), continuity, states that small variations in the components a, b of the interval [a,b] cannot alter the final result drastically.

**A4.** Continuity of * over the interior of region S

The next axiom (A5) deals with the concept of identity.

**A5.** The interval [0,1] is the identity,

$$[a,b] * [0,1] = [0,1] * [a,b] = [a,b] \quad (2)$$

The motivation for this axiom is that when an evidence [a,b] with some definite information is combined with another evidence $[\varepsilon_1, 1-\varepsilon_2]$ where $\varepsilon_1$ and $\varepsilon_2$ are very small (i.e. this evidence has very little information content), then the final interval must be close to [a,b], the evidence with substantial information.

However, the combination of [a,b] and $[\varepsilon_1, 1-\varepsilon_2]$ with interval Bayesian decision rule will yield very conservative result, i.e., the final interval will be close to $[\varepsilon_1, 1-\varepsilon_2]$ ([Cheng & Kashyap, 1988]). Interval Bayesian rule doesn't satisfy (A5) and hence won't be discussed. The final necessary property or axiom is symmetry (A6).

**A6.** Symmetry of an interval

If $[a,b] * [c,d] = [e,f]$, then

$$[1-b, 1-a] * [1-d, 1-c] = [1-f, 1-e] \quad (3)$$

Recall that [a,b] means that the basic probability in support of H, varies from a to b. But this also means that basic probability in support of $\overline{H}$ varies from (1-b) to (1-a).

We also need the enhancement property when the component evidences are not in conflict. This will be discussed later.

## 3. Conflicting Evidence

Recall that an interval [a,b] regarding a hypothesis H means that support for H varies from a to b and the support for $\overline{H}$, the negation of H varies from (1-b) to (1-a). Let

Discrimination

$\triangleq$ (lower limit of support of H) −

(lower limit of support of $\overline{H}$)

$$\triangleq a-(1-b)=a+b-1 \qquad (4)$$

If the discrimination is positive, we will regard H as true, if forced to make a decision and $\overline{H}$ is true if discrimination is negative. Thus two evidences are in conflict if their discrimination measures are of opposite signs. We will state this as a definition.

*Definition* (conflicting evidences). Two evidences specified by intervals [a,b] and [c,d] are said to be in conflict if (a+b-1) and (c+d-1) are of opposite signs.

A pair of evidences which do not obey the above definition is said to be non-conflicting. Geometrically, the pair [a,b] and [c,d] are in conflict if vectors (a,b) and (c,d) do not fall in the same triangle BCD and ACD in Figure 1.

It should be noted that a pair of intervals of [a,b] and [c,d] may be overlapping and still in conflict, say [0.1, 0.6] and [0.4, 0.9]. The intuitive idea is that most of the interval of the first evidence falls in [0, 0.5] (or [0.5, 1]) and the most of the second interval in [0.5, 1] (or respectively [0, 0.5]).

Recall that evidence with interval [0,1] has no information in it or it has maximum measure of imprecision. We can regard the width of an interval as measure of imprecision. Suppose we have a pair of *conflicting* evidences [a,b] and [c,d] with width (b-a) and (d-c) respectively. Then we expect the measure of imprecision of the final result, say [e,f], be greater than the width of the individual components, i.e.,

(B1) If [a,b] and [c,d] are conflicting, and [a,b]*[c,d]=[e,f] then the combining rule * is reasonable if $|f-e| \geq \max[|b-a|,|d-c|]$.

When two evidences are *not* conflicting, then both the evidences support H (or $\overline{H}$) (not both), and we expect the combining rule to have reinforcing property in (B2):

(B2) When two intervals [a,b] and [c,d] are in the same subset [0.5, 1] (or [0, 0.5]), the combination rule should possess the following reinforcing property

$$[a,b] * [c,d] = [e,f]$$

$$|f-e| \leq \min[|b-a|,|d-c|]$$

i.e., narrower the width, greater is our measure precision regarding the final result.

Experimental scientists like astronomers have routinely use this reinforcing feature, i.e., if two different experiments give intervals [0.6, 0.8] and [0.7, 0.9], then the combined interval should support hypothesis strongly, i.e., the width of result be less than of its components.

## 4. A New Approach for Handling Conflicting Evidences

We need a fresh strategy to construct combination rules which obey both the necessary axioms (A1)-(A6), the conflict resolution property (B1) and the reinforcement property (B2) in non-conflict situations. The strategy is to reinforce the strength of non-conflicting portion of evidence and decrease the strength of conflicting portion. We can envision the following three steps:

**Step 1:** Homeomorphically transform (a,b), (a,b)∈ S, into (u,v), (u,v)∈ R [Mostert & Shields, 1957], so that the one component u is equivalent to the discrimination measure (i.e., support for H minus support for $\overline{H}$ given in (4)). u < 0 (>0) means discrimination is negative (positive) respectively.

**Step 2:** Let $(u_1,v_1)$ and $(u_2,v_2)$ be the maps of

$(a_1,b_1)$ and $(a_2,b_2)$, the two intervals. Combine $(u_1,v_1)$ and $(u_2,v_2)$ by any function which preserves associativity and commutativity ([Bonissone, 1987][Cheng & Kashyap, 1989][Hajek, 1985]). Let

$$u = f_1(u_1, u_2) \qquad (5)$$
$$v = f_2(v_1, v_2)$$

The function $f_1$ plays the key role in combining conflicting and non-conflicting components. $f_2$ affects the width of interval.

**Step 3:** Map $(u,v) \in R$ back into the corresponding point in S, say $(e,f) \in S$; the resulting interval is $[e,f]$.

The step 1 assures the satisfaction of the closure axiom (A1), the symmetry axiom (A6) and the conflict resolution property. Step 2 assures the satisfaction of the remaining properties. We will give one specific rule:

**T-P combination rule** (Triangle to plane map)

**Step 1.** Here we map the permissible triangle region, ABC, into a plane region as shown in Figure 2. Note $c=(0,1) \rightarrow c'=(0,0)$. The point $(a_1, b_1)$ is mapped to $(u_1, v_1)$:

$$u_1 = \frac{(a_1+b_1-1)}{a_1^2+(1-b_1)^2} * \sum_{i=0}^{N}(1+a_1-b_1)^{i+2}, \qquad (6)$$

$$v_1 = \frac{2a_1(1-b_1)}{a_1^2+(1-b_1)^2} * \sum_{i=0}^{N}(1+a_1-b_1)^{i+1}$$

Note as $N \rightarrow \infty$, $\sum_{i=0}^{N}(1+a_1-b_1)^i = 1/(b_1-a_1)$. If $(a_1+b_1-1)<0$, then $u_1<0$ and if $(a_1+b_1-1)>0$, then $u_1>0$. Hence if a pair of evidences $(a_1,b_1)$ and $(a_2,b_2)$ are conflicting, then their corresponding u's component will be of opposite signs.

**Step 2.** If $[a_1, b_1]$ is mapped to $[u_1, v_1]$ and $[a_2, b_2]$ is mapped to $[u_2, v_2]$, then we will use a simple addition rule to combine them

$$u = u_1 + u_2 \qquad (7)$$
$$v = v_1 + v_2$$

**Step 3.** Invert the $(u,v)$ coordinates to triangle region giving the final result:

$$[e,f] = [t\frac{1}{1+\tan(\frac{\theta}{2})}, \; 1-t\frac{\tan(\frac{\theta}{2})}{1+\tan(\frac{\theta}{2})}], \qquad (8)$$

where

$$\theta = \cos^{-1}\frac{u_1+u_2}{\sqrt{(u_1+u_2)^2+(v_1+v_2)^2}}$$

$$t = \frac{\sqrt{(u_1+u_2)^2+(v_1+v_2)^2}}{1+\sqrt{(u_1+u_2)^2+(v_1+v_2)^2}}$$

The width of the resulting interval is:

$$|f-e| = 1-t = \frac{1}{1+\sqrt{(u_1+u_2)^2+(v_1+v_2)^2}} \qquad (9)$$

The TP rule satisfies (B1) when the two evidences are highly conflicting, i.e., $|u_1| \gg v_1$, $|u_2| \gg v_2$, $|u_1| \approx |u_2|$ and $u_1$ and $u_2$ are of opposite signs (v ≥ 0 always), hence $(u_1+u_2)^2+(v_1+v_2)^2 \ll u_1^2$ or $u_2^2$. Hence r is smaller, i.e., the width $|f-e|$ increases. TP rule satisfies (B2) automatically, because if the two evidences are not conflicting, i.e., $u_1$ and $u_2$ are of the same sign, r is greater than both $\sqrt{u_1^2+v_1^2}$ and $\sqrt{u_2^2+v_2^2}$ and consequently, by (9) the width $|f-e|$ decreases.

## 5. Numerical Computation with Conflicting Evidences

EXAMPLE 1. Take the diagnosis of the severity of jaundice of a patient. Suppose the patient is checked by two doctors. The report from the first doctor shows the condition of ja ndice is slight, whereas the report from the second doctor indicates the condition of jaundice is severe. A conclusion such as "the condition of jaundice of patient A is moderate" is hardly acceptable. It is more appropriate to remain indecisive.

As a numerical illustration, let interval [0,0] (or [1,1]) denote that the severity of jaundice is definitely slight (or severe), respectively. Suppose the assessment of severity of jaundice according to the first test, T1, and the second test, T2, is

T1 : [0.2, 0.4]
T2 : [0.7, 0.9]

The two judgments are conflicting since T1 and T2 are contained in two disparate halfs, [0,0.5] and [0.5,1], respectively. The severity of patient A's jaundice given by different rules are

D-S : [0.57, 0.64]
T-R : [0.42, 0.65]

As mentioned earlier, the interval given by DS has width 0.07 which is much less than the intervals in the original evidence namely 0.2. Looking at DS result, there is no indication that it is obtained from two conflicting evidences. The entire interval [0.57, 0.64] lies in the [0.5,1] indicating an acceptance of the hypothesis of jaundice which is completely unacceptable. The interval given by TR is acceptable.

EXAMPLE 2. Consider several pair of evidences which are not conflicting.

| data | D-S | T-P |
|---|---|---|
| [0.1, 0.2]*[0.3, 0.4] | [0.1, 0.11] | [0.22, 0.28] |
| [0.2, 0.6]*[0.2, 0.6] | [0.24, 0.43] | [0.25, 0.5] |
| [0, 0.3]*[0, 0.4] | [0, 0.12] | [0, 0.21] |

Here the DS's property of reinforcement all-the-time is handy. TR satisfies the property of reinforcement in these data sets.

### 6. Dependency Handling

Recall that when two experts arrive at intervals use the same raw data, combining the evidences by D-S rule is roughly equivalent to counting the same evidence twice. How do we develop a combining operation when a pair of evidences are partially dependent? The modified TP rule will be stated.

- **mTP rule**

Recall that the domain of (u,v) coordinates of TP rule derived earlier is the upper half plane. We need the family of cT(x,y,p) functions ([Cheng & Kashyap, 1989]) to modify the TP rule:

$$cT(x,y,p) \quad (10)$$

$$=(x^p + y^p)^{\frac{1}{p}}, \quad x,y \geq 0$$

$$=-((-x)^p + (-y)^p)^{\frac{1}{p}}, \quad x,y \leq 0$$

$$=(-(-x)^p + y^p)^{\frac{1}{p}}, \quad x \leq 0 \leq y \ \& \ (-x) \leq y$$

$$=-((-x)^p - y^p)^{\frac{1}{p}}, \quad x \leq 0 \leq y \ \& \ (-x) \geq y$$

Using (a,b)→(u,v) coordinates transformation function (6) and substituting (24) for the addition operation (7) in (8) yields the following *mTP* rule

$$[e,f] = [t\frac{1}{1+\tan(\frac{\theta}{2})}, \ 1-t\frac{\tan(\frac{\theta}{2})}{1+\tan(\frac{\theta}{2})}], \quad (11)$$

where

$$\theta = \cos^{-1}\frac{cT(u_1,u_2,p)}{\sqrt{cT(u_1,u_2,p)^2 + cT(v_1,v_2,p)^2}}$$

$$t = \frac{\sqrt{cT(u_1,u_2,p)^2 + cT(v_1,v_2,p)^2}}{1+\sqrt{cT(u_1,u_2,p)^2 + cT(v_1,v_2,p)^2}}$$

The TP rule corresponds to p=1, the independent case. As p→∞, $cT(u_1,u_2,p)$ and $cT(v_1,v_2,p)$ tend to $Max[u_1,u_2]$ and $Max[v_1,v_2]$ which describe the extreme case of dependency between two evidences in (u,v) coordinates. From a control viewpoint, if probabilistic relation about two evidences $e_1$ and $e_2$ is given, i.e., $\alpha_1 = p(e_2|e_1)$ and $\alpha_2 = p(e_1|e_2)$, the value of p is determined by the measure shown in (12)

$$p = \frac{(\alpha_1+\alpha_2)}{2-(\alpha_1+\alpha_2)} \quad (12)$$

Roughly speaking, the overall contribution from two independent evidences to the final decision is assumed to be greater than that from a pair of dependent evidences.

## 7. Numerical Comparison for Rules of Dependency Handling

The following examples will show the combination with highly dependent evidences by DS, TP and mTP (eq.(11)) rules.

EXAMPLE 3. [Kyburg, 1987] Let's assume that at least 70% of the soft berries ($e_1$) in a certain area are good to eat, and that at least 60% of the red berries ($e_2$) are good to eat. The dependency measure between soft and red berries is given to be 0.9 (soft berries are red) and 0.7 (red berries are soft). What are the chances that a soft red berry is good to eat? Dempster's rule yields [0.6, 1]*[0.7, 1]=[0.88, 1] and TP rule yields [0.6, 1]*[0.7, 1]=[0.79, 1]. Both rules show the enhancement in belief that a berry is good to eat if it is soft and red. However, the result is unduly optimistic since the two attributes are related. By taking the dependency between two attributes, red and soft, into account with p=4 (from (12)), mTP rule yields [0.6, 1]*[0.7, 1]=[0.71, 1]. This result indicates relatively small amount of increase in belief. This appears to be much better. □

EXAMPLE 4. Estimates of whether an approaching airplane is a warplane or a commercial jetliner are reported by two passive radars. The dependency between the two radars can be viewed as a function of the angle between radar sites and the target where the target is the vertex. Suppose that the airplane approaches in a direction which makes the angle small so that there is a large amount of overlapping in detection and hence the two estimates are considered to be highly dependent. Suppose the value of p is estimated to be 10. Supports of the approaching airplane being a warplane provided by two radars are:

radar 1 :   [0.6,0.8]
radar 2 :   [0.7,0.9]

The combined result should not be very different from original intervals since two reports are highly dependent. Results obtained from D-S, T-P and mTP (p=10) rules are

D-S :   [0.85, 0.9]
T-P :   [0.71, 0.82]
mTP :   [0.65, 0.82]

The D-S and T-P rules all show an increase of support in the hypothesis "approaching airplane being a warplane" since the lower limit of supports, 0.85 from D-S, 0.71 from T-P are greater than lower limits of original intervals. As expected, T-P and D-S rules gives a too optimistic estimate. Whereas the value given by rule mTP shows much smoother increase in support. □

## 8. Conclusion

We have discussed the problem of evidence combination and proposed a method for computing weights of support, conflict and ignorance by using a geometrical model. Based on the proposed approach, we could obtain combining operations which provide acceptable results in the conflicting cases as well as dependent evidences. The proposed combining rules possess several properties which often are not taken into account by other approaches in evidence combination. We have compared combining rules of D-S and TP with different types of evidences. Results show that the proposed TP rule may increase the width when evidences are conflicting, and like D-S rule decrease the width when they are not. The suggested construction method also allows one to modify the TP rule with cT function to cope with dependent evidences without loss of those fundamental properties. Several applications show that the modified TP rule provide results as expected when evidences are dependent.

### Acknowledgement
The authors would like to thank Dr. Lashon B. Booker for his helpful comments.

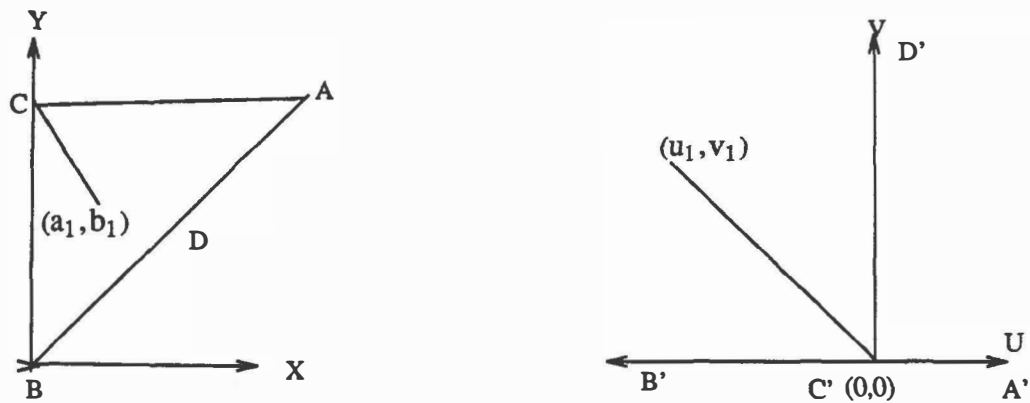

**Figure 2** Triangle to Plane Map